\documentclass{svproc}
\usepackage{epsfig}
\usepackage{graphicx}
\usepackage{amsmath}
\usepackage{amssymb}
\usepackage{subcaption}
\usepackage{placeins}

\usepackage{url}

\begin{document}
\mainmatter              
%
\title{Seam Carving Detection and Localization using Two-Stage Deep Neural Networks}
\titlerunning{Seam Carving Detection}  
%
\author{Lakshmanan Nataraj \and Chandrakanth Gudavalli \and
Tajuddin Manhar Mohammed \and Shivkumar Chandrasekaran \and B.S. Manjunath }
\authorrunning{Lakshmanan Nataraj et al.} 
%
\tocauthor{Lakshmanan Nataraj, Chandrakanth Gudavalli, Tajuddin Manhar Mohammed, Shivkumar Chandrasekaran, and B.S. Manjunath}
\institute{Mayachitra Inc., Santa Barbara, CA, USA,\\
\texttt{https://mayachitra.com/}
}

\maketitle              

\begin{abstract}
Seam carving is a method to resize an image in a content aware fashion. However, this method can also be used to carve out objects from images.
In this paper, we propose a two-step method to detect and localize seam carved images. 
First, we build a detector to detect small patches in an image that has been seam carved.
Next, we compute a heatmap on an image based on the patch detector's output. 
Using these heatmaps, we build another detector to detect if a whole image is seam carved or not.
Our experimental results show that our approach is effective in detecting and localizing seam carved images.
\end{abstract}

\section{Introduction}

\begin{figure}[t]
\centering
\includegraphics[width=0.9\linewidth]{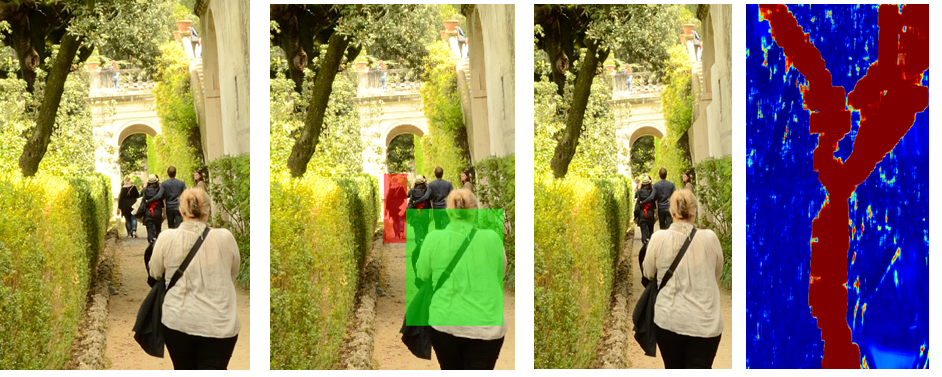}
\vspace{-5pt}
\caption{Illustration of seam carving detection and localization: (a) Original image, (b) Object marked in red to be removed and Object marked in green to be preserved, (c) Seam carved image with object removed, (d) Seam carving detection heatmap using proposed approach (red pixels are areas where seams were likely removed). }
\label{fig:seam-carve-det-eg}
\vspace{-10pt}
\end{figure}

With new cameras, mobile phones and digital tablets, the amount of digital images has had an exponential increase.
Social media platforms have also contributed to their increased distribution.
At the same time, software for manipulating these digital images have also significantly evolved.
These software tools make it trivial for people to manipulate these digital images.
The objective of Media Forensics is to identify these manipulations and detect these doctored images.
Over the years, many techniques have been proposed to identify image manipulations.
These include digital artificats based on camera forensics, resampling characteristics, compression, and others.
A common operation in image tampering is removing certain image regions in a ``content-aware'' way.
In this regard, seam carving is a popular technique for ``content-aware'' image resizing~\cite{seam_carving_avidan,shamir2009seam}
In seam carving, the ``important content'' in an image is left unaffected when the image is resized and it is generally assumed that the ``important content'' is not characterized by the low energy pixels.
Since seam carving based object removal involves non-traditional ways of removing objects, it is a challenge to detect doctored images that have been seam carved.
In this paper, we propose a novel method to detect and localize seam carved images using two stages of convolutional neural networks (CNNs): one for detection and one for localization. 
First, we train a CNN to identify patches that have been seam carved.
For every pixel in an image, we then compute the detection score which results in a heatmap for the whole image, that can be used for localization.
Finally, we train another CNN on the heatmaps which gives a score at the image level to determine if an image has been seam carved or not.
Fig.~\ref{fig:seam-carve-det-eg} illustrates the proposed approach. 



The rest of the paper is organized as follows.
Sec.~\ref{sec:relwork} presents the related work in this area and Sec.~\ref{sec:seam-car} introduces seam carving and seam insertion on images.
The methodology to detect seam carving is presented in Sec.~\ref{sec:method} while the experiments are detailed in Sec.~\ref{sec:exps}.
Finally, the conclusion is presented in Sec.~\ref{sec:conc}.

\FloatBarrier
\section{Related Work}
\label{sec:relwork}

There have been several works proposed to detect digital image manipulations. 
These include detection of splicing, morphing, resampling artifacts, copy-move, seam carving, computer-generated (CG) images, JPEG artifacts, inpainting, compression artifacts, to name a few.
Many methods have been proposed to detect copy-move~\cite{li2015segmentation,cozzolino2015efficient}, resampling~\cite{popescu-farid-resampling,babak-radon,kirchner-local,Nataraj10-345,ryu2014estimation,feng2012normalized,bayar2017robustness,bunk2017detection}, splicing~\cite{guillemot2014image,bappy2019hybrid,salloum2018image}, and inpainting based object removal~\cite{wu2008detection,liang2015efficient}. 
Other approaches exploit JPEG compression artifacts~\cite{farid2009exposing,lin2009fast,luo2010jpeg,bianchi2011improved} or artifacts arising from artificial intelligence (AI) generated images~\cite{marra2018detection,zhang2019detecting,goebel2020detection,barni2020cnn}.
In recent years, deep learning based methods have shown better performance in detecting image manipulations~\cite{bayar2016deep,bayar2017design,rao2016deep,bunk2017detection}.

Several methods have been proposed over the past decade to detect seam carving based manipulations~\cite{sarkar2009detection,lu2011seam,liu2013improved,chang2013detection,wei2013patch,liu2014improved,wattanachote2015tamper,fei2015detection,sheng2016detection,zhang2017detection,sheng2017detection,han2018exploring,gong2018detecting,zhang2020detecting,li2020identification}. 
These include methods using steganalysis~\cite{sarkar2009detection}, hashing~\cite{lu2011seam,fei2015detection}, local binary pattern~\cite{zhang2017detection,zhang2020detecting}, and deep learning based methods~\cite{ye2018convolutional,cieslak2018seam,nam2019content,nam2020deep}. 
In this paper, our approach to detect seam carving based manipulations is also based on deep learning.



\begin{figure}[t]
\centering
\includegraphics[height=0.4\columnwidth]{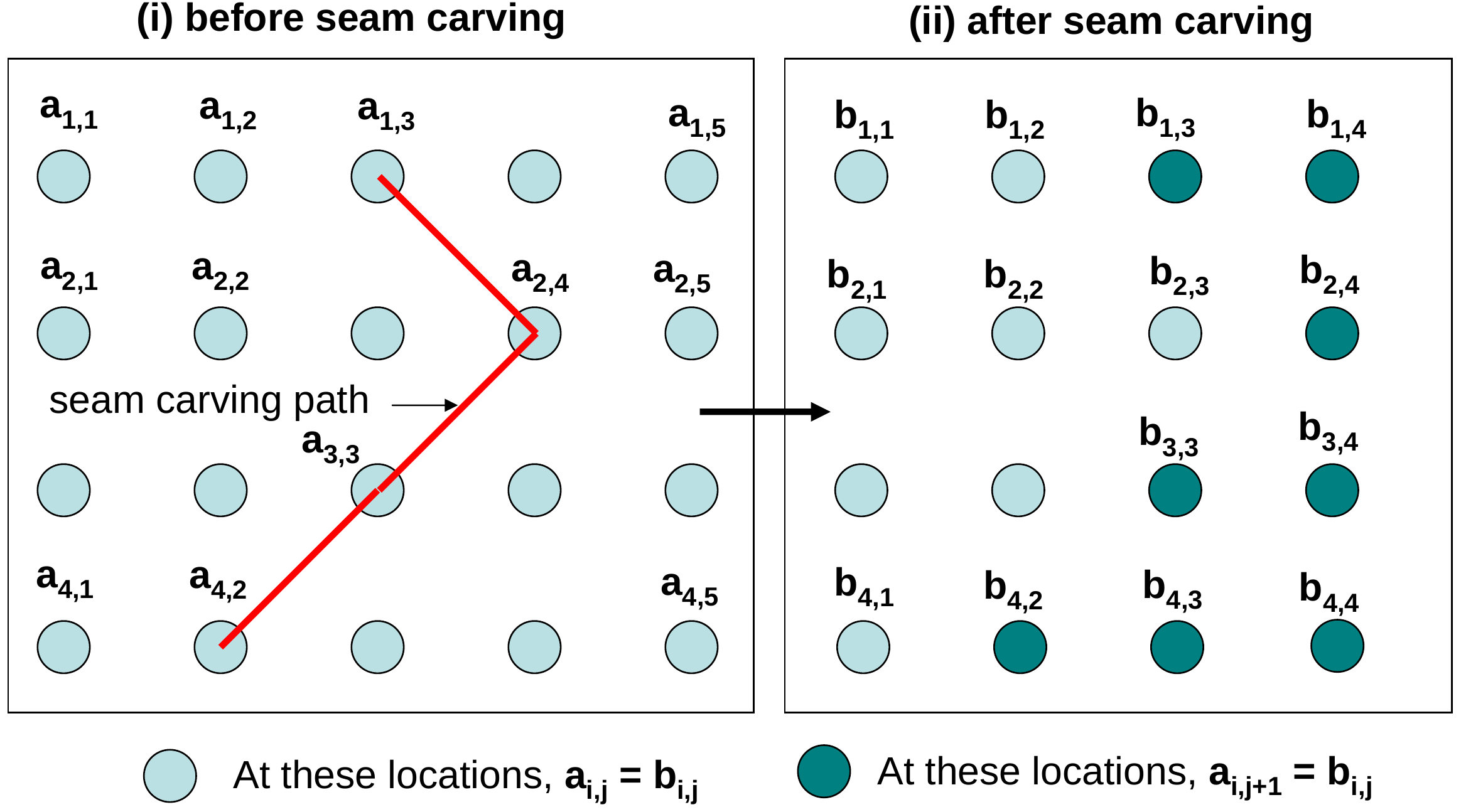}
\vspace{-5pt}
\caption{Example of Seam Carving when a $4 \times 5$ matrix ${\bf a}$ is seam carved and a $4 \times 4$ matrix ${\bf b}$ results due to the removal of a single seam.}
\label{fig:explain-seam-carving-01}
\vspace{-10pt}
\end{figure}

\begin{figure}[b]
\centering
\includegraphics[height=0.32\columnwidth]{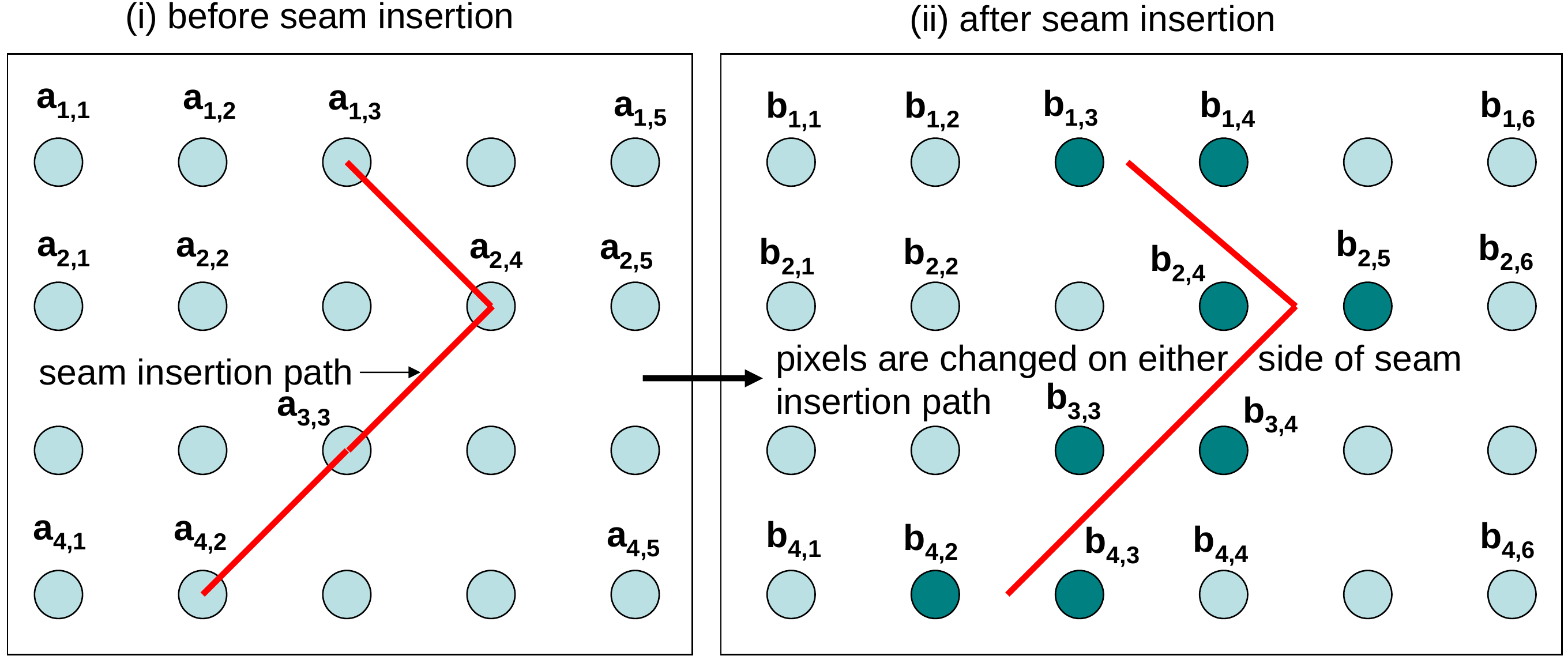}
\vspace{-5pt}
\caption{Example of Seam Insertion: (i) ${\bf a}$ and (ii) ${\bf b}$ are the $4 \times 5$ and the $4 \times 6$ image matrices before and after seam insertion, respectively. For points along the seam, the values are modified as shown for the first row: $b_{1,1}\! = \! a_{1,1}, ~ b_{1,2}\! = \! a_{1,2}, ~ b_{1,3} \! = \!\mbox{round}(\frac{a_{1,2} + a_{1,3}}{2}), ~b_{1,4} \! = \!\mbox{round}(\frac{a_{1,3} + a_{1,4}}{2}), ~b_{1,5}\!= \! a_{1,4},~ b_{1,6} \! = \! a_{1,5}$.}
\label{fig:explain-seam-carving-02}
\vspace{-10pt}
\end{figure}

\section{Seam Carving and Seam Insertion}
\label{sec:seam-car}

A seam is defined as an optimal 8-connected path of pixels on an image either from top-to-bottom or left-to-right.
In seam carving, the seams are removed and the image dimension is reduced by a column or a row.
In seam insertion, a seam is first removed and two pixels are inserted at the position where the seam was removed.
Fig.~\ref{fig:explain-seam-carving-01} and \ref{fig:explain-seam-carving-02} illustrates the processes of seam carving and seam insertion.
An energy function computed for all points along a seam is considered for the optimality criterion for seam selection.
This choice of seams helps in maintaining the image quality during the resizing process.
{\it We consider the seam carved/inserted image as a tampered image because the image dimensions and it's content are altered.}.
Hence, the problem of detecting seam carving/insertion is important from an image forensics perspective.
Interpolation kernel based methods for re-sampling detection will fail when the resizing in the doctored image is done using seam carving/insertion.
Though it was initially proposed for automatic image resizing while maintaining a good perceptual quality of the resized image, seam carving has also been used for removal of certain image regions. {\it It is to be noted that seam carving can discard and retain certain regions, depending on the weight we assign to certain regions}. E.g. for an object removal problem, we may need to ensure that certain image regions are left unaffected as distorting them may cause significant perceptual distortion. We first explain how seam carving is used for object removal and then discuss the interesting problems involved.

\begin{figure}[t]
  \centering
  \begin{subfigure}{.84\textwidth}
  \centering
  \includegraphics[width=0.9\linewidth]{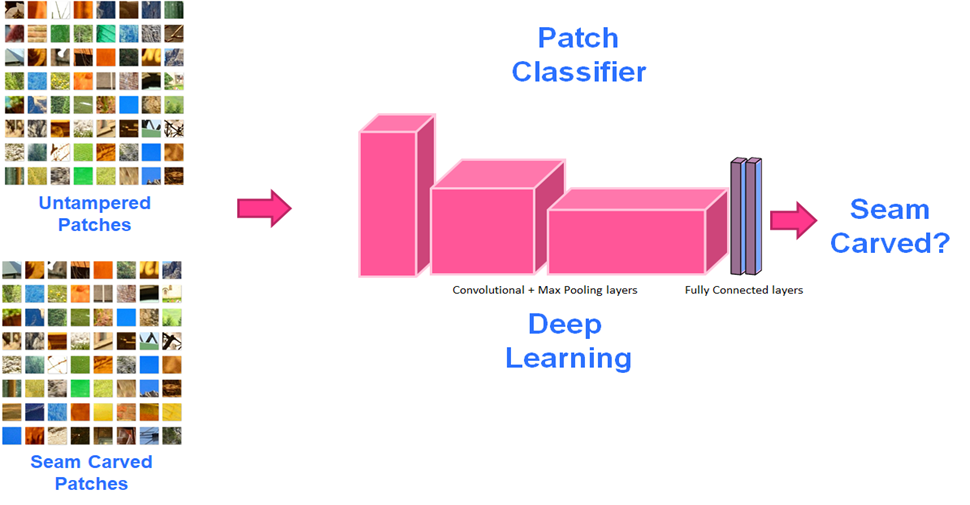}
  \caption{Stage 1}
\end{subfigure}\\%
\begin{subfigure}{.84\textwidth}
  \centering
  \includegraphics[width=0.9\linewidth]{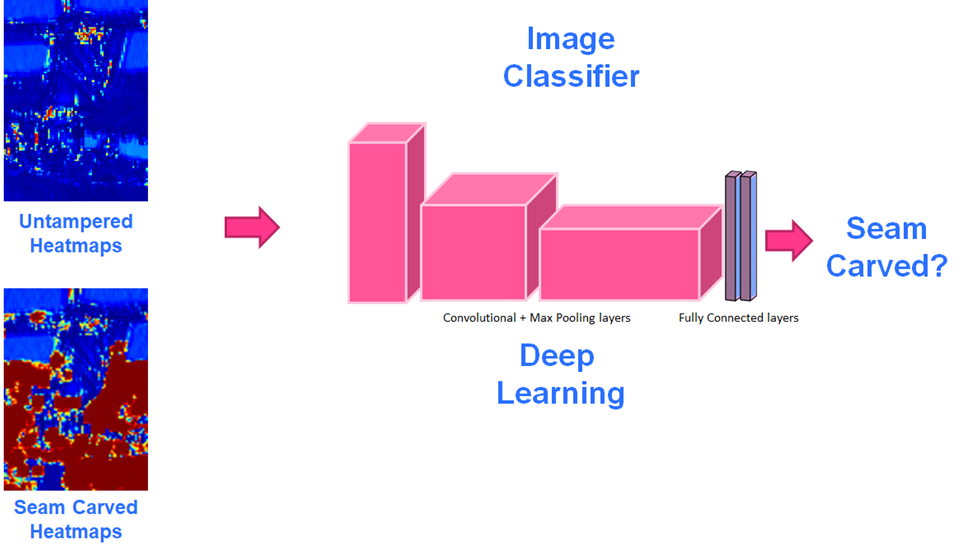}
  \caption{Stage 2}
\end{subfigure}
\vspace{-5pt}
\caption{Two-stage approach}
\label{fig:seam-carve-detn-approach}
\end{figure}

\section{Detection of Seam Carving}
\label{sec:method}

\begin{figure}[t]
\centering
\includegraphics[width=0.44\linewidth]{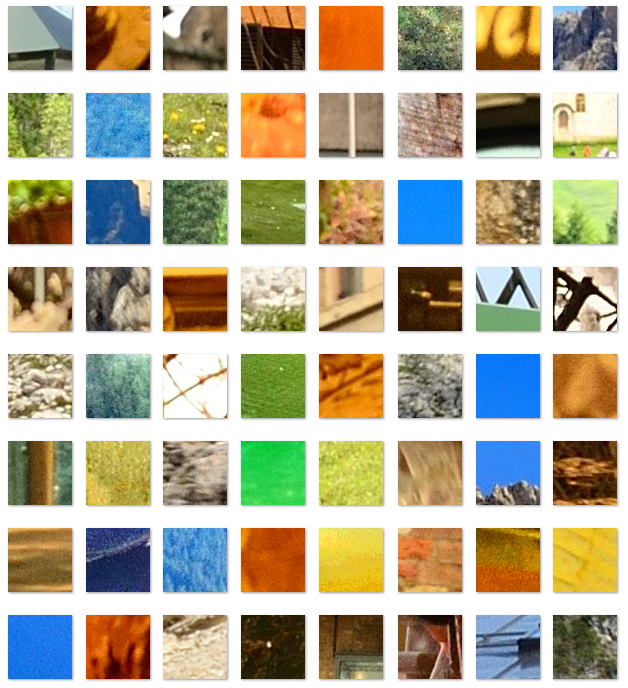} \ \ \ \ 
\includegraphics[width=0.44\linewidth]{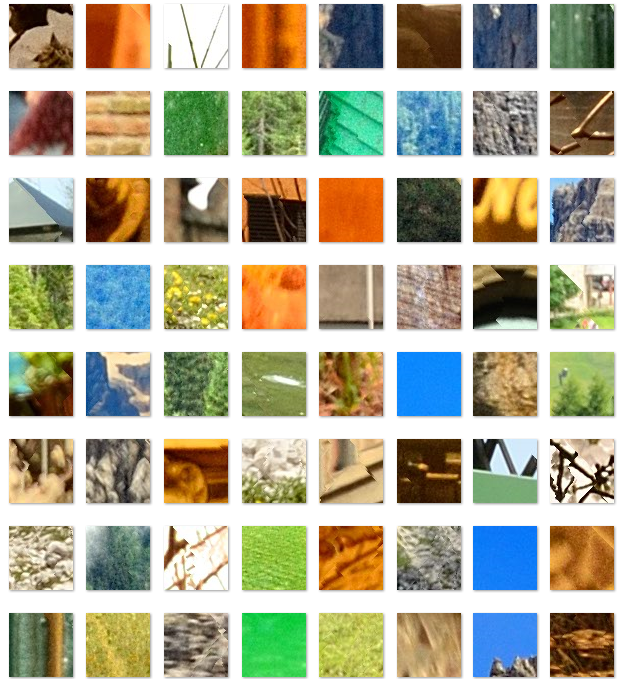}
\vspace{-5pt}
\caption{Screenshot of: (a) Non-seam carved patches, (b) Seam carved patches} 
\label{fig:scr-patches}
\vspace{-10pt} 
\end{figure}

In order to detect and localize seam carving in images, we propose a two-stage detection approach: one for detection of seam carved patches and the other for localizing seam carving in an image by generating a heatmap.
First, we train a deep neural network to identify whether patches in an image have been seam carved or not. 
We then divide an image into patches and for every patch, we compute the detection score which results in a heatmap for the whole image.
This heatmap can be used for localization of seam carving.
Finally, we train another deep neural network with the heatmaps as input which gives a score at the image level to determine whether an image has been seam carved or not.
The entire block schematic is shown in Fig.~\ref{fig:seam-carve-detn-approach}


\section{Experiments}
\label{sec:exps}
\subsection{Experimental Setup} 

We first extract $64 \times 128$ patches from images belonging to RAISE dataset~\cite{dang2015raise}.
From these patches, we form two classes of image patches: first class where the patches are further cropped to $64 \times 64$, and the second class where the patches are seam carved horizontally by 50\% to obtain $64 \times 64$ seam carved patches. 
In this way we obtained 16,000 patches from the RAISE dataset (8000 in each class) and 40,000 patches from the Dresden dataset (20,000 in each class).
These were further randomly divided into 80\% training, 10\% testing and 10\% validation.  


\subsection{Learning}
\label{sec:learn}

The patches are trained using a multi layer deep convolutional neural network which consists of convolution layer with 32 3x3 convs, followed by ReLu layer, convolution layer with 32 5x5 convs followed by max pooling layer,  convolution layer with 64 3x3 convs followed by ReLu layer, convolution layer with 64 5x5 convs followed by max pooling layer, convolution layer with 128 3x3 convs followed by ReLu layer, convolution layer with 128 5x5 convs followed by max pooling layer, and finally a 256 dense layer followed by a 256 dense layer and a sigmoid layer. 
We train this model till a high training accuracy and validation accuracy are obtained.

\subsection{Detection Heatmaps}

\begin{figure}[t]
\centering
\includegraphics[width=0.42\linewidth,height=0.7\linewidth]{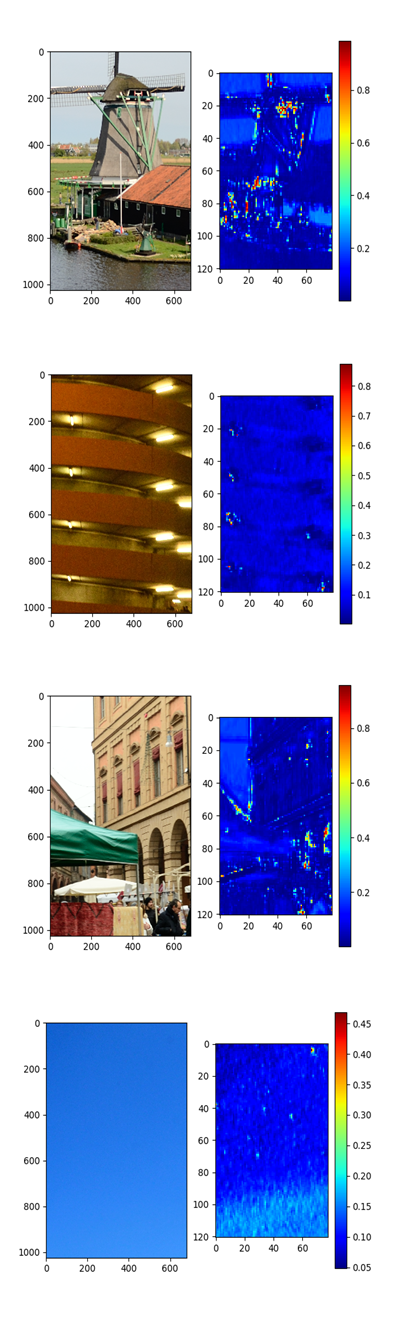} \ \ \ \ 
\includegraphics[width=0.42\linewidth,height=0.7\linewidth]{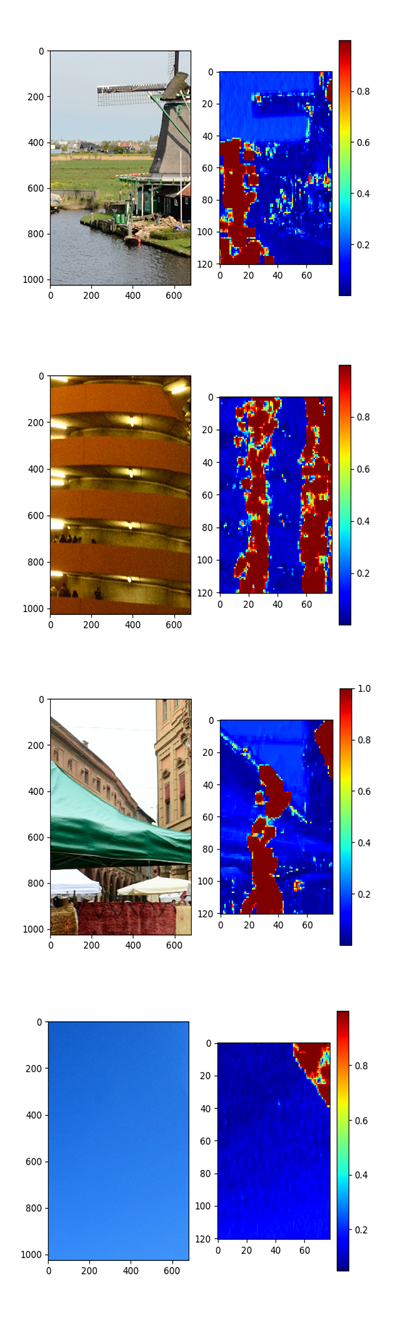}
\vspace{-5pt}
\caption{Detection heatmaps on images that have (a) not been seam carved, and (b) seam carved} 
\label{fig:hmaps-sc}
\vspace{-10pt} 
\end{figure}

\begin{figure}[h]
\centering
\includegraphics[width=0.6\linewidth]{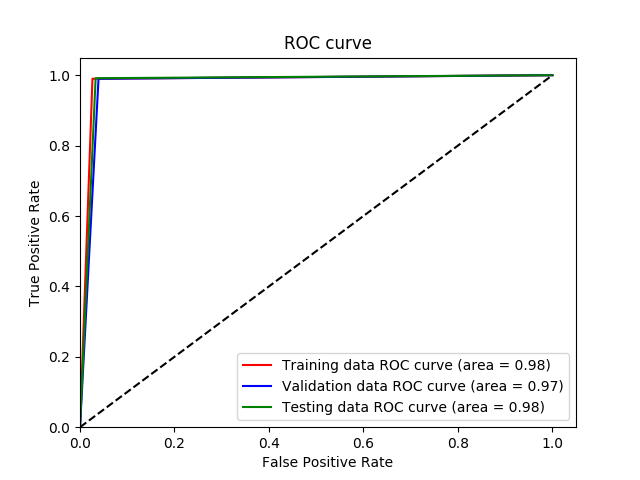}
\vspace{-5pt}
\caption{ROC curve of seam carving detection on the model trained on the heatmaps} 
\label{fig:raise-64-heatmap}
\vspace{-10pt} 
\end{figure}

Using the trained model on the patches, the probability of a pixel being seam carved or not is computed on overlapping patches in an image.
Fig.~\ref{fig:hmaps-sc} show the heatmaps on non-seam carved and seam carved images.
As we can see, the heatmaps on the seam carved images have more red regions than the images on non-seam carved images.
Even for an image that has the blue sky, the heatmaps can be clearly identified for seam carved image and the non-seam carved image. 
This motivated us to train the heatmaps with another CNN which takes the heatmaps as input (Fig.~\ref{fig:seam-carve-detn-approach}(b)) and outputs the probability whether an image has been seam carved or not. 
As we can see from Fig.~\ref{fig:raise-64-heatmap}, we obtained high accuracy when trained on the heatmaps.

\vspace{-5pt}
\begin{table}[t]
\caption{Robustness to Percentage of Seams Removed}
\vspace{+2pt}
\centering
\begin{tabular}{|l|c|}
\hline
Percentage & Area under the curve (AUC) \\
\hline
\hspace{15pt} 1 & 0.6464 \\ \hline
\hspace{15pt} 2 & 0.7838 \\ \hline
\hspace{15pt} 5 & 0.9274 \\ \hline
\hspace{15pt} 8 & 0.9540 \\ \hline
\hspace{12pt} 10 & 0.9724 \\ \hline
\hspace{12pt} 20  &     0.9866 \\ \hline
\hspace{12pt} 30   &   \textbf{0.9919} \\ \hline
\hspace{12pt} 40   &   \textbf{0.9937} \\ \hline
\hspace{12pt} 50   & \textbf{0.9916} \\ \hline
\hspace{12pt} 60   & 0.9502 \\ \hline
\hspace{12pt} 70   &   0.9150 \\ \hline
\hspace{12pt} 80   & 0.8670 \\ \hline
\hspace{12pt} 90   & 0.8223 \\ 
\hline
\end{tabular}
\label{tab:perc-seams-rev} 
\end{table}

\subsection{Robustness to Percentage of seams removed}

In this experiment, we varied the percentage of seams removed in the testing set and evaluated the model which was trained with 50\% seams removed, in order to check the robustness of the model for different amounts of seams removed.
The Area Under the Curve (AUC) is the evaluation metric for varying percentage of seams removed.
The results are tabulated in Tab.~\ref{tab:perc-seams-rev}.
We observe that the AUC is very high for percentages around 50\% and decreases for lower percentages of seams removed.
This shows that the model is generalizable for most percentages of seams removed. 
In future, we will train another model for lower percentages.


\begin{figure}[h]
\centering
\includegraphics[width=0.95\columnwidth]{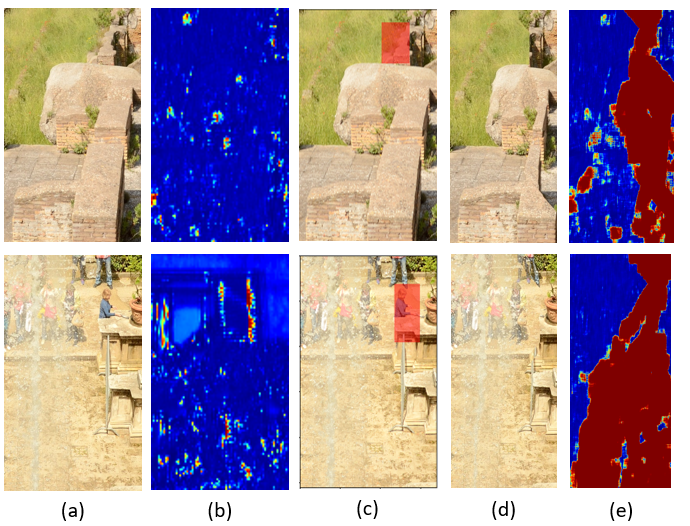}
\vspace{-5pt}
\caption{Detection heatmaps on images where objects have been removed using seam carving: (a) original image, (b) heatmap computed on original image, (c) object marked for removal in red, (d) image with object removed using seam carving, (e) heatmap computed on object removed image showing the possible seam paths}
\label{fig:obj-rem-1}
\vspace{-10pt}
\end{figure}

\begin{figure}[t]
\centering
\includegraphics[width=0.95\columnwidth]{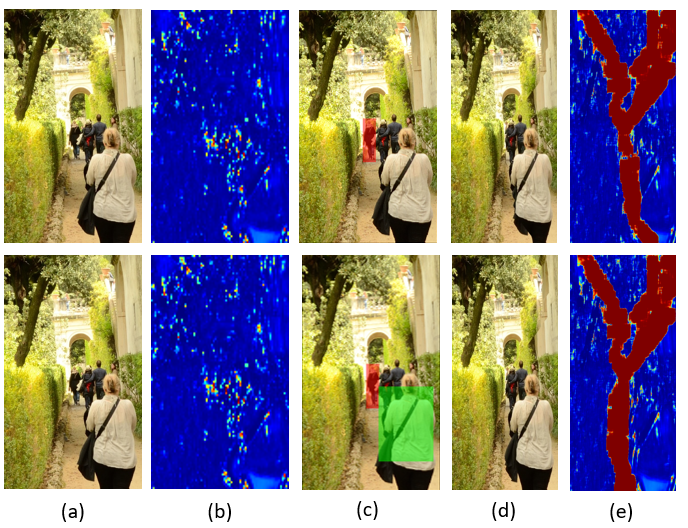}
\vspace{-5pt}
\caption{Explainability in the heatmaps: (a) original image, (b) heatmap computed on original image, (c) object marked for removal in red and area preserved in green (bottom row) (d) image with object removed using seam carving (in the top row - person's leg is removed while preserved in the bottom row), (e) heatmap computed on object removed image showing the possible seam paths with explainability. The seam paths change on the top row and bottom row near the person's leg.}
\label{fig:obj-rem-exp}
\vspace{-10pt}
\end{figure}

\subsection{Robustness to JPEG compression}
In this experiment, we evaluated the robustness of our proposed against JPEG compression. We varied the JPEG Quality Factors (QFs) of test images from 100 to 50. 
The model was trained on seam carved and non-seam carved patches and images, which were also JPEG compressed between the Quality Factors of 70-100. 
The Area Under the Curve (AUC) is chosen as the evaluation metric.
The results are tabulated in Tab.~\ref{tab:jpeg-exp}.
We observe that the AUC is high when the QF is high (compression is low) and the AUC reduces as the QF decreases (compression increases). 
However, even at a QF of 50, the AUC is still reasonably high.  

\begin{table}[b]
\caption{Robustness to JPEG Compression}
\centering
\begin{tabular}{|l|c|}
\hline
JPEG Quality Factor (QF) & Area under the curve (AUC) \\
\hline
\hspace{46pt} 100 & 0.9376 \\ \hline
\hspace{48pt} 90 & 0.9160 \\ \hline
\hspace{48pt} 80 & 0.8578 \\ \hline
\hspace{48pt} 70 & 0.8027 \\ \hline
\hspace{48pt} 60 & 0.7658 \\ \hline
\hspace{48pt} 50 & 0.7332 \\ \hline
\end{tabular}
\label{tab:jpeg-exp} 
\end{table}


\subsection{Explainability on Object Removed Images}

Here, we evaluate our approach in a practical scenario where objects are removed in images using seam carving. 
We chose an object or a region in an image that has to be removed. 
The weights of this region are set to a low value such that the seam carving algorithm is forced to pass through this region, thus removing the object from the image.
When our approach was evaluated on these images, we observe that our model is able to localize the region that was removed as well as the paths taken by the seam carving algorithm as shown in Fig.~\ref{fig:obj-rem-1}.

The detection heatmaps also exhibit explainability as shown in Fig.~\ref{fig:obj-rem-exp} where an object is marked for removal in red. 
While this object is removed successfully, a person's leg in the foreground also gets removed (top row). 
To prevent this, another area is marked in green (bottom row) by giving high weights so that the person's legs are not removed.
As we can see from the heatmaps computed on the seam carved images (top and bottom row), the path showing the possible seams also changes near the person's legs, thus exhibiting explainability.


\subsection{Extension to Seam Insertion Detection}

\begin{figure}[t]
\centering
\includegraphics[width=0.6\linewidth]{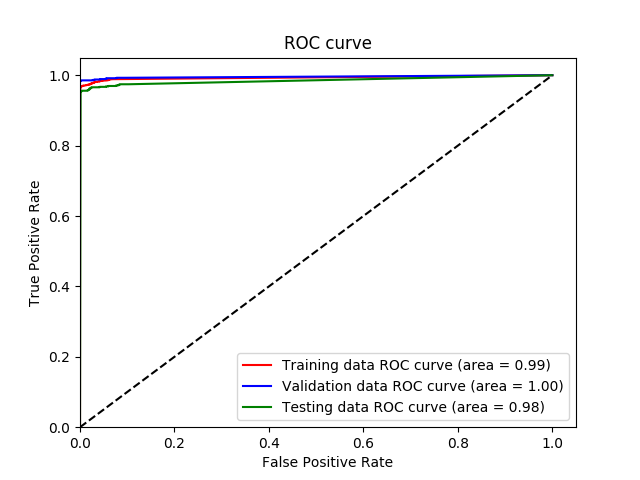}
\vspace{-5pt}
\caption{ROC curves at image level to detect seam inserted images}
\label{fig:hmaps-im-si}
\vspace{-10pt}
\end{figure}

Finally, we also extend the seam carving detection approach to detecting seam insertion. 
We first extract $64 \times 64$ patches from images belonging to RAISE dataset~\cite{dang2015raise}. 
From these patches, we form two classes of image patches: first class where the patches are further cropped to $64 \times 64$, and the second class where the patches are seam inserted from $64 \times 32$ dimensions to $64 \times 64$ seam dimensions. 
In this way we obtained 16,000 patches from the RAISE dataset (8000 in each class). 
These were further randomly divided into 80\% training, 10\% testing and 10\% validation.  
The patches are trained using a multi layer convolutional neural network as explained in Sec.~\ref{sec:learn}.
We train this model till a high training accuracy and validation accuracy are obtained.
Using the trained model on the patches, the probability of a pixel being seam inserted or not is computed on overlapping patches in an image to produce a heatmap. 
Another model is trained on the heatmaps to determine if an image has seam insertions or not.
As we can see from Fig.~\ref{fig:hmaps-im-si}, we obtained high accuracy when trained on the heatmaps.

\section{Conclusion and Future Work}
\label{sec:conc}

In this paper, we presented an approach to detect seam carved images.
Using two stages of CNNs, we detect and localize areas in an image that have been seam carved.
In future, we will focus on making our detections more robust, combining seam carving and insertions, and also extend to other object removal methods such as inpainting.

\section{Acknowledgements}
This research was developed with funding from the Defense Advanced Research Projects Agency (DARPA).
The views, opinions and/or findings expressed are those of the author and should not be interpreted as representing the official views or policies of the Department of Defense or the U.S. Government. 
The paper is approved for public release, distribution unlimited.

\FloatBarrier


%

%
%








\bibliographystyle{splncs03} 
\bibliography{ucr,forensics-mc,forensics_citations}

\end{document}